\documentclass[letterpaper, 10 pt, conference]{IEEEtran}
\IEEEoverridecommandlockouts    

\usepackage{multirow}
\usepackage{lipsum}
\usepackage{amsmath}
\usepackage{amssymb}
\usepackage[ruled,vlined]{algorithm2e}
\usepackage{graphicx}
\usepackage{amsmath,amssymb,amsfonts}
\usepackage{algorithmic}
\usepackage{graphicx}
\usepackage{textcomp}
\usepackage{xcolor}
\usepackage{graphicx}
\usepackage{multirow}
\usepackage{threeparttable}   
\usepackage{makecell}         
\usepackage{multirow}  
\usepackage{xcolor}
\usepackage{threeparttablex}
\graphicspath{{./Figures/}}

\title{\LARGE \bf Beyond Known Objects: A Novel Framework for Open-Set Object Detection using Negative-Aware Norm}

\author{
	\parbox{\textwidth}{%
		\centering
		Yuchen Zhang$^{*}$, Yao Lu$^{*}$, Johannes Betz%
	}%
    \thanks{$^{*}$These authors contributed equally to this work.}%
	\thanks{Y. Zhang, Y. Lu and J. Betz are with the Professorship of Autonomous Vehicle Systems, TUM School of Engineering and Design, Technical University of Munich, 85748 Garching, Germany; Munich Institute of Robotics and Machine Intelligence (MIRMI). {\tt\small yuchen2.zhang@tum.de}}%
}

\usepackage{cite}
\usepackage{hyperref}

\begin{document}
	
	\maketitle
    \definecolor{known}{HTML}{fcb515}  
    \definecolor{unknown}{HTML}{0381ff}  
    \definecolor{darkred}{HTML}{93110e}  
    \definecolor{background}{HTML}{e5d4e8}  
	\thispagestyle{empty}
	\pagestyle{empty}
	
	\begin{abstract}
		Open-Set Object Detection (OSOD) is crucial for autonomous driving, where perception systems must recognize and localize both known and previously unseen objects in complex, dynamic environments. While recent approaches deliver promising results, they often require retraining the detector extensively to learn objectness, which describes the likelihood that a bounding box tightly encloses a valid object, regardless of whether its category was learned during training. Deviating from existing work, we hypothesize that standard off-the-shelf detectors may already contain helpful cues for objectness, owing to their training on numerous and diverse known categories. Building on this idea, we propose NAN-SPOT, a training-light framework that does not require to retrain the base object detector and estimates objectness by leveraging a hidden layer metric called Negative-Aware Norm (NAN), requiring only minutes of training on just hundreds of images. To support comprehensive evaluation, we introduce COCO-Open, an expanded version of the existing COCO-Mixed dataset, increasing unknown object annotations from 433 to 1853, making it the most exhaustively labeled dataset for OSOD to the best of our knowledge. Experimental results demonstrate that NAN-SPOT achieves even better performance on unknown object detection than methods requiring heavy training, without compromising performance on known objects. This efficiency and robustness make NAN-SPOT a promising step towards open-world perception in autonomous driving. The code and the dataset will be made publicly available.
	\end{abstract}
	
	\section{Introduction}
	\label{sec:introduction}

    Object detection is a core component of perception systems in intelligent vehicles, supporting safety-critical functions such as obstacle perception, risk assessment, and motion planning. Despite substantial progress, most deployed object detectors \cite{varghese2024yolov8, centernet, bevformer} are based on a closed-set formulation and are trained and evaluated on a fixed set of predefined semantic categories, including vehicles, pedestrians, and cyclists. While these detectors perform well under controlled settings, they remain limited to recognizing object categories observed during training and lack explicit mechanisms for handling unknown objects in open and dynamic traffic environments \cite{bogdoll2022anomaly}. In real-world driving scenarios, vehicles inevitably encounter previously unseen obstacles, such as road debris, fallen cargo, or temporary construction materials, that are absent from training annotations. When such objects fall outside the predefined label set, conventional detectors may fail to recognize them, leading to an underestimation of environmental risk at the perception level and preventing downstream planning modules from taking appropriate evasive or conservative actions \cite{gao2025foundation}. To compensate for these perceptual uncertainties, deployed systems often rely on sensor redundancy and conservative fallback strategies, for example by triggering an emergency stop, rather than explicitly modeling unknown objects at the perception stage \cite{saad2022perceptual}.

    \begin{figure}[t]
    \begin{center}
    \includegraphics[width=\linewidth]{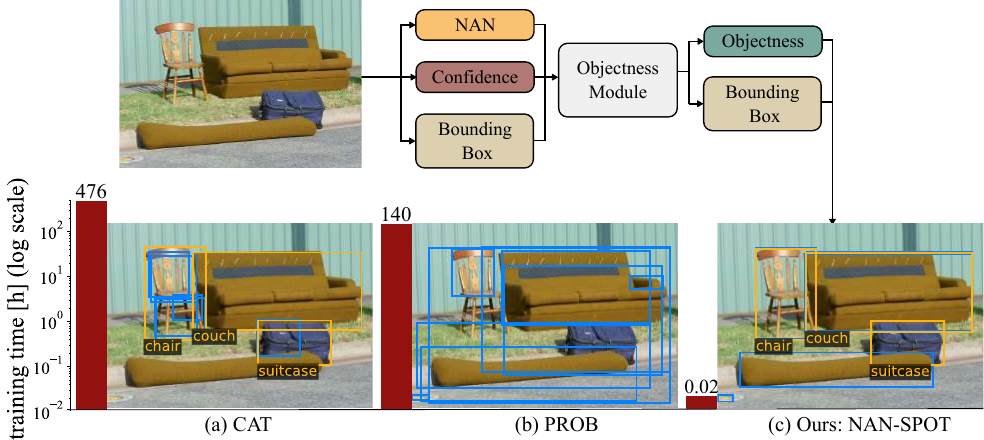} 
    \caption{\textbf{Top}: NAN-SPOT extends D-DETR with an objectness module estimating objectness scores for each bounding box, supported by the NAN metric derived from the detector’s final hidden layer. \textbf{Bottom}: Predicted \textcolor{unknown}{unknown}/ \textcolor{known}{known} objects and training time of CAT \cite{ma2023cat}, PROB \cite{zohar2023prob} and our model. }
    \label{fig:front}
    \end{center}
    \end{figure}

    To address this limitation, recent research has introduced Open-Set Object Detection (OSOD) \cite{du2022wild, fontanel2022unknown, cheng2024uadet, dhamija2020overlooked, liang2023unknown, li2023novel, kong2021opengan} that extends the detection task to open-world settings, requiring models not only to detect and classify known objects but also to identify previously unknown objects from the background. This dual requirement imposes two key challenges: (1) the objectness objective, which demands reliable separation of both known and unknown objects from background, and (2) the classification objective, which ensures accurate labeling of known classes while flagging unknown instances without misclassifications. %
    Most existing approaches \cite{joseph2021towards,gupta2022owdetr,ma2023cat,doan2024hyp} tackle these challenges by using pseudo-labeling and require extensive retraining. %
    
    In this work, we propose an alternative approach based on the Negative-Aware Norm (NAN) \cite{park2023understanding}, a metric originally designed for out-of-distribution (OOD) detection for image classification. We present a training-light framework built on Deformable DEtection TRansformer (D-DETR) and adopt the NAN metric for objectness estimation. We refer to this framework as NAN-SPOT. By leveraging NAN calculated from the output of the last decoder layer, we introduce a lightweight class-agnostic objectness estimator capable of distinguishing both known and unknown objects from background. Trained on just 504 images from COCO-OOD \cite{liang2023unknown}, our method achieves performance that exceeds methods requiring extensive training, with an example detection shown in Figure~\ref{fig:front}.

    Furthermore, to support rigorous OSOD evaluation, we introduce COCO-Open, an enhanced version of COCO-Mixed dataset \cite{liang2023unknown}, where we manually expand the annotation for both known and unknown objects. In particular, we provide over four times as many unknown object annotations as the original. This improved labeling allows for more reliable and comprehensive assessment of the detection of unknown objects.
    Our key contributions can be summarized as follows: 
    \begin{itemize}
        \item We present NAN-SPOT, a training-light framework for OSOD that leverages the NAN metric to model objectness, enabling effective separation of both known and unknown objects from background.%
        \item We release COCO-Open, a dataset with comprehensive known and unknown object annotations for evaluating OSOD and related tasks.
    \end{itemize}

    \section{Related Work}
    \subsection{End-to-end Object Detectors}
    DEtection TRansformer (DETR)~\cite{carion2020end} introduced a transformer-based architecture for object detection that replaced traditional hand-designed components like anchor boxes and non-maximum suppression with a set-based bipartite matching. While offering a clean end-to-end formulation, DETR converges slowly and struggles with small objects due to its global attention mechanism. To address these limitations, D-DETR~\cite{zhu2020deformable} incorporated sparse, multi-scale deformable attention, enabling faster training and improved performance on small objects. As a result, D-DETR has become a widely adopted backbone for a variety of vision tasks \cite{han2024idpd, shanliang2022airport, ma2023cat, chen2024accurate, bar2022detreg, zhang2023dino}. Our work also builds upon D-DETR, leveraging its strengths while introducing an objectness estimation module to further enhance its ability to detect unknown objects in open-world settings.

    \subsection{Open-Set Object Detection}
    OSOD focuses on detecting unknown objects during inference while maintaining accurate classification of known categories. Open-World Object Detection (OWOD) extends this paradigm by integrating incremental learning, allowing models to gradually incorporate unknown objects as known once the labels are provided, while avoiding catastrophic forgetting. Both settings share the core challenge of identifying unknown objects reliably. 
    
    Several methods have been proposed to tackle this problem. UnSniffer \cite{liang2023unknown} developed a generalized object confidence score to highlight object regions by training on proposals that partially or fully overlapped with annotated known objects, leveraging a combination of regression and contrastive losses. Subsequently, an energy score was introduced to distinguish known objects from unknown ones.

    ORE \cite{joseph2021towards} combined energy-based scoring for unknown detection with contrastive clustering to structure the embedding space, enhancing discrimination of unknowns within the Faster R-CNN \cite{ren2016faster} framework. Sun et al. \cite{sun2024exploring} proposed disentangling the objectness estimation task and the classification task by representing them as the magnitude and orientation in a polar coordinate system, respectively. Transformer-based architectures were also adapted to OSOD. OW-DETR \cite{gupta2022owdetr}, building on D-DETR, introduced attention-guided pseudo-labeling to guide the learning of unknown objects. CAT \cite{ma2023cat} enhanced detection through cascade decoding and adaptive label refinement, improving both localization and classification in open-world settings. PROB \cite{zohar2023prob} modeled objectness probabilistically using a class-agnostic Gaussian distribution, which significantly boosted recall on unknown objects. Finally, Hyp-OW \cite{doan2024hyp} projects detection query embeddings into hyperbolic space and enforces hierarchical structure among known classes via contrastive learning with a superclass regularizer. Hyperbolic distance from known-class centroids is utilized to adaptively pseudo-label unmatched detections as unknowns.

    While these methods demonstrate strong performance, they often require extensive training time and complex pipelines to achieve competitive results. In contrast, our approach achieves better results with drastically reduced training time, requiring only a few minutes.

	\section{Method}
	\label{sec:Method}

    \subsection{Problem Formulation}
    During training, the model is given \( N \) labeled images \( \{\mathbf{I}_i, \mathbf{Y}_i\} \), where each annotation \( \mathbf{y}_k = [c_k, \mathbf{b}_k] \) includes a known class label $ c_k \in \mathcal{K} = \{1,2,\dots,C\}$ and a bounding box $\mathbf{b_k}$. At inference time, the model operates in an open-world setting, where it must not only detect and classify objects from the known categories $\mathcal{K}$, but also identify and localize novel objects from unknown classes \( \mathcal{U} = \{C+1, C+2, \ldots\} \).

\begin{figure*}[htbp]
    \begin{center}
    \includegraphics[width=0.65\linewidth]{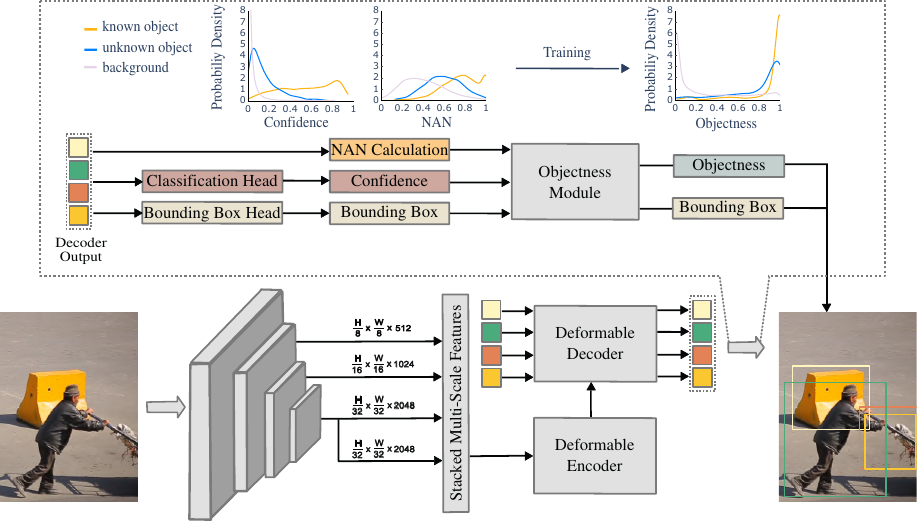} 
    \end{center}
       \caption{\textbf{Overview of the proposed NAN-SPOT for open-set object detection. (Bottom)} Based on D-DETR, which extracts multi-scale feature maps from a CNN backbone and encodes them using a transformer encoder with deformable attention to focus on sparse, informative regions. A set of learnable query embeddings is then passed to the decoder, where they are iteratively refined via cross-attention with the encoded features.
       \textbf{(Top) Objectness estimation:} The decoder output supports the classification and bounding box heads and is additionally used to compute the NAN metric. These three components are subsequently fed into the objectness estimation module. The resulting objectness score effectively separates both known and unknown objects from background, which is evidenced by the probability density curves shown before and after training. This separation enables more reliable detection of foreground objects.}
    \label{fig:pipeline}
    \end{figure*}

    To address this challenge, we introduce NAN-SPOT, a training-light framework for OSOD. Its overall pipeline is illustrated in Fig.~\ref{fig:pipeline} and detailed in Section~\ref{sec:NAN-SPOT}. Furthermore, we present COCO-Open, a dataset with exhaustive unknown object annotations for OSOD, described in Section~\ref{sec:coco-open}.
    
    \subsection{NAN-SPOT}
    \label{sec:NAN-SPOT}

    \subsubsection{Negative-Aware Norm (NAN)}
    \label{sec:NAN}
    The norm of hidden layer feature vectors has long been used to distinguish in-distribution (ID) from OOD samples \cite{norm1, norm2, norm3, norm4}, with higher norms typically indicating ID instances. However, Park et al.~\cite{park2023understanding} showed that this metric alone is unreliable, as it overlooks neuron deactivation patterns that can cause ID samples to be misclassified as OOD. To address this, they proposed NAN, which combines activation sparsity (i.e., the count of inactive neurons) with vector norm to improve OOD detection robustness. Formally, given a latent space vector \( \mathbf{z} = [z_1, z_2, \dots, z_d]^\top \in \mathbb{R}^d \), NAN is defined as:
    \begin{equation}
        \|\mathbf{z}\|_{\text{NAN}} = \|\mathbf{z}\|_1 \cdot \|\mathbf{z}\|_0^{-1}, \label{eqn:nan_main}
    \end{equation}
    where \( \|\mathbf{z}\|_1 \) is the \(\ell_1\) norm,
    \begin{equation}
        \|\mathbf{z}\|_1 = \sum_{i=1}^{d} |z_i|, \label{eqn:nan_a1}
    \end{equation}
    and \( \|\mathbf{z}\|_0 \) represents the number of activated neurons,
    \begin{equation}
        \|\mathbf{z}\|_0 = d - |\{i : z_i \leq 0 \}|. \label{eqn:nan_a0}
    \end{equation}

    While NAN has proven effective in distinguishing ID from OOD samples in image classification, its applicability to object detection remains unclear. 
    Unlike image classification, where each image is assumed to contain either a single known or an unknown object, object detection presents a more nuanced challenge. Each predicted bounding box may correspond to one of three general cases: (1) a single known object, (2) a single unknown object, or (3) all other regions that the model should ignore--such as background, multiple objects or poorly localized object--which we refer to collectively as background throughout this paper.
    This raises a key question: \textbf{When applied to object detection, does NAN favor known objects over unknowns and background, or does it instead capture a more general notion of objectness, distinguishing both known and unknown objects from the background?}
    
    \subsubsection{A Proof of Concept}
\begin{figure*}
    \centering
    \begin{minipage}{0.65\textwidth}
        \centering
        \includegraphics[width=0.8\linewidth]{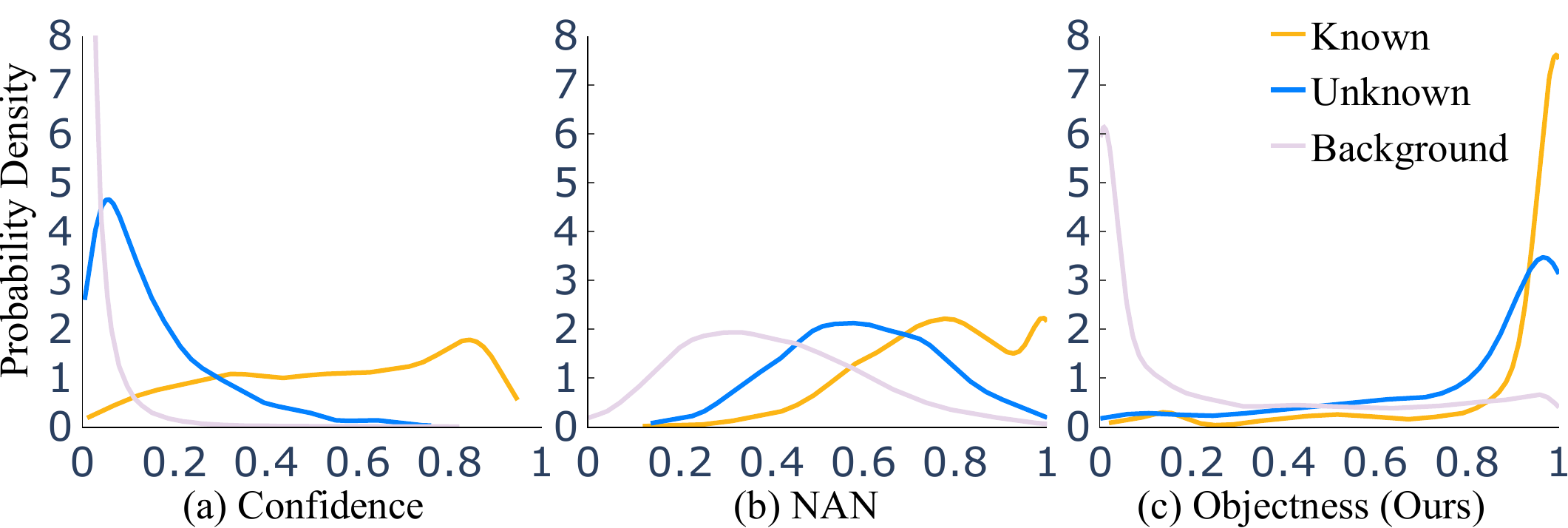}
    \end{minipage}
    \begin{minipage}{0.28\textwidth}
        \centering
        \resizebox{\linewidth}{!}{
        \begin{tabular}{|l|c|c|c|}
        \hline
        Metric & \makecell{Bg-Kn \\($\uparrow$)} & \makecell{Bg-Unk \\($\uparrow$)}  & \makecell{Unk-Kn \\($\downarrow$)}  \\
        \hline\hline
        Confidence          & 0.530 & 0.126 & 0.405 \\
        NAN                 & 0.366 & 0.195 & 0.172 \\
        Objectness   & \textbf{0.648} & \textbf{0.538} & \textbf{0.109}\\
        \hline
        \end{tabular}}
    \end{minipage}
     \caption{
        \textbf{A proof-of-concept for D-DETR metrics on COCO-Open. (left) Distribution of metrics across \textcolor{known}{known} /\textcolor{unknown}{unknown} objects and \textcolor{background}{background} regions.} KDE is applied to estimate the probability density. \textbf{(a)} Confidence shows substantial overlap between unknowns and background. \textbf{(b)} NAN metric assign higher scores to unknown but introduces confusion for background. \textbf{(c)} Objectness shows clear separation, with known and unknown objects concentrated at high values and background at lower end.
        \textbf{(right) Separation quality measured by Wasserstein-1 distance. } Objectness achieves the best background separation while minimizing the gap between known and unknown, providing a more robust notion of objectness. Abbreviations: Bg = Background, Kn = Known, Unk = Unknown, Obj = Object. Best results are in \textbf{bold}.
        }
    \label{fig:kdes}
    \end{figure*}

    To investigate this problem, we analyze two scoring functions: the NAN and the conventional confidence score, both of which can be extracted from an off-the-shelf D-DETR model~\cite{zhu2020deformable}. The confidence score is the maximum predicted probability over known classes, $p_{\text{conf}} = \max_{c \in \mathcal{C}_{\text{known}}} h_{\text{cls}}^{(c)}(\mathbf{f})$, where $h_{\text{cls}}(\mathbf{f})$ is the classification head applied to the decoder output $\mathbf{f}$. NAN is computed on the last decoder output as $f_{\text{NAN}} = \|\mathbf{f}\|_{\text{NAN}}$, where $\|\cdot\|_{\text{NAN}}$ is defined in Eq. \eqref{eqn:nan_main}. %
    An ideal objectness measure should produce high separation between background and objects (both known and unknown), while keeping the distributions of known and unknown objects close. %
    In Fig.~\ref{fig:kdes}, the visualization of distribution of both scores are produced with kernel density estimation (KDE) on the COCO-Open dataset. Separability is quantified using pairwise Wasserstein-1 distance. It measures the minimal cost of transporting probability mass to align two distributions, providing a principled notion of separation.

\begin{figure}[t]
    \begin{center}
    \includegraphics[width=0.75\linewidth]{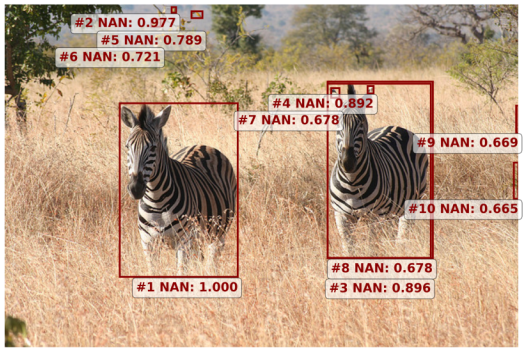} 
    \end{center}
       \caption{\textbf{Top 10 detections by NAN.} While successfully assigning high scores to salient objects (e.g., both zebras), NAN also consistently assigns high scores to tiny regions and bounding boxes along the edges of the image. 
       }
    \label{fig:topnan}
    \end{figure}

    As shown in Fig.~\ref{fig:kdes}, confidence succeeds in separating known objects from background but fails to distinguish unknowns, which largely overlap with background. NAN alleviates this issue by improving background-unknown separation (Bg-Unk: $0.126$ to $0.195$) and assigning higher values to unknown objects, reducing the gap between known and unknown (Unk-Kn: $0.405$ to $0.172$). These observations highlight that confidence and NAN capture complementary aspects of detection: confidence suppresses background well but overlooks unknowns, while NAN responds stronger to unknowns but remains sensitive to background artifacts, i.e., NAN may produce elevated scores in non-object regions such as small fragments or edges (an example can be seen in Fig.~\ref{fig:topnan}). This complementarity motivates the joint use of confidence and NAN, while suggesting that incorporating box-level cues could further mitigate spurious responses and strengthen localization.

    \subsubsection{Objectness Estimation}  
    Building on these observations, we introduce an objectness estimator $g(\cdot)$ that integrates multiple complementary cues. Specifically, the input features consist of (i) the NAN score $f_{\text{NAN}}$, (ii) classification confidence $p_{\text{conf}}$, and (iii) box-related features $\mathbf{b} = [\,s_{\text{box}}, d_{\text{center}}, d_{\text{edge}}\,]$, where $s_{\text{box}}$ is the box size, $d_{\text{center}}$ the minimum distance of the box center to any image edge, and $d_{\text{edge}}$ the minimum distance from any box edge to the image boundary. The estimator predicts the likelihood that a detection corresponds to any valid object, whether known or unknown.

    The estimator is designed as a lightweight module that can be appended to a pretrained D-DETR, which implicitly learns to provide class agnostic region proposals \cite{carion2020end, zhou2023unknown}. It requires only minimal additional supervision in the form of a small annotated dataset where both known and unknown objects are labeled, which is significantly less data than required for full detector retraining. %
    In principle, $g(\cdot)$ may be realized using any classification method. In our experiments, we evaluate two lightweight estimators: a random forest \cite{breiman2001random} and a multilayer perceptron (MLP) \cite{rumelhart1986learning}. Both are chosen for their efficiency and effectiveness. %
    The resulting objectness score is defined as
    \begin{equation}
    p_{\text{obj}} = g(\mathbf{f}) = g\big(f_{\text{NAN}}, \; p_{\text{conf}}, \; \mathbf{b}\big).
    \end{equation}
    The objectness $p_{\text{obj}}$ provides a more discriminative measure for object presence. Quantitative and qualitative evaluation in Fig.~\ref{fig:kdes} demonstrates clear improvements: our objectness achieves superior separation between background and objects (Bg-Kn: $0.648$, Bg-Unk: $0.538$) while maintaining minimal distance between known and unknown objects (Unk-Kn: $0.109$). This results in a more consistent and reliable measure of objectness suitable for open-set object detection scenarios. 
    
    \subsection{COCO-Open Dataset}
    \label{sec:coco-open}
\begin{figure}[htbp]
    \centering
    
    \resizebox{0.9\linewidth}{!}{%
    \begin{tabular}{cc}
    \includegraphics[width=0.40\linewidth]{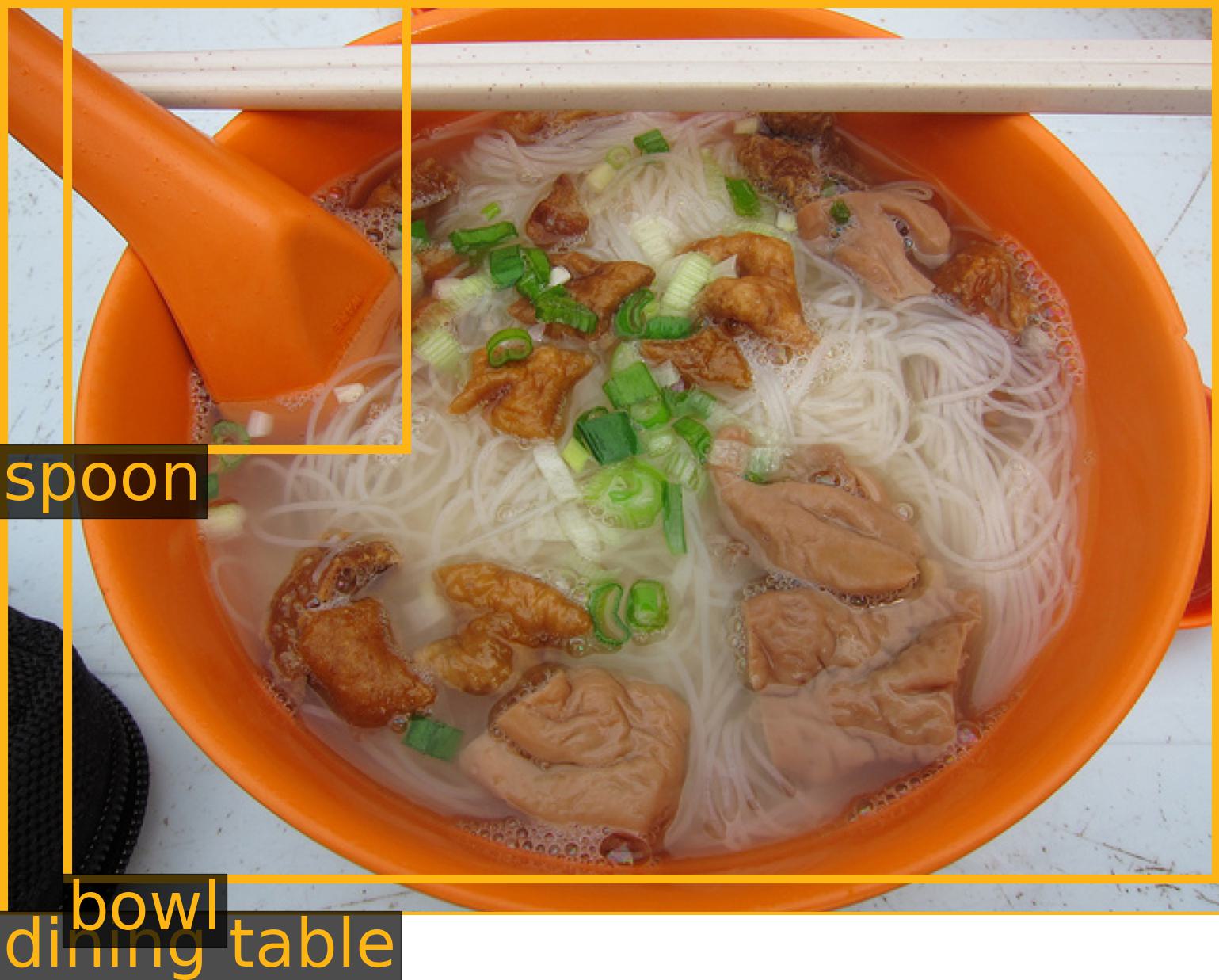} &
    \includegraphics[width=0.23\linewidth]{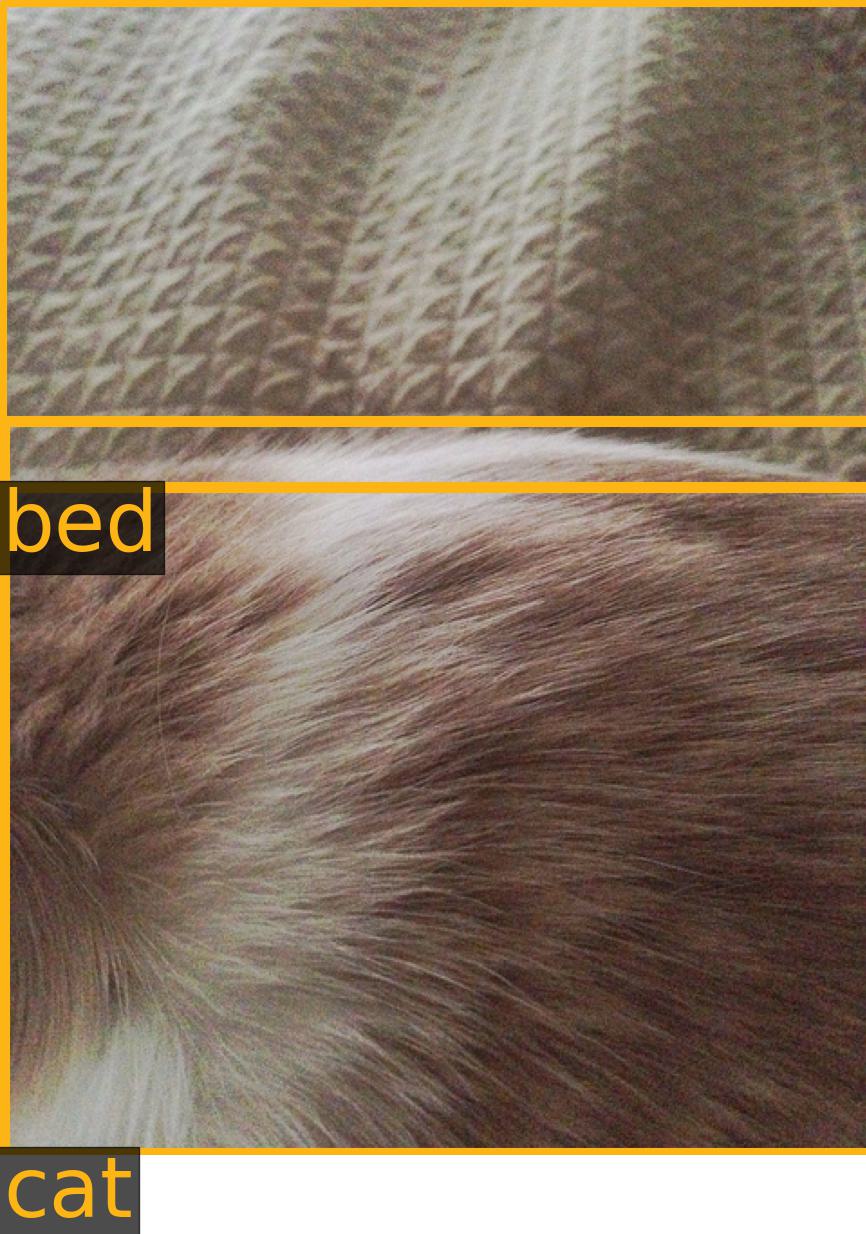} \\
    (a) & (b) \\[4pt]
    \includegraphics[width=0.48\linewidth]{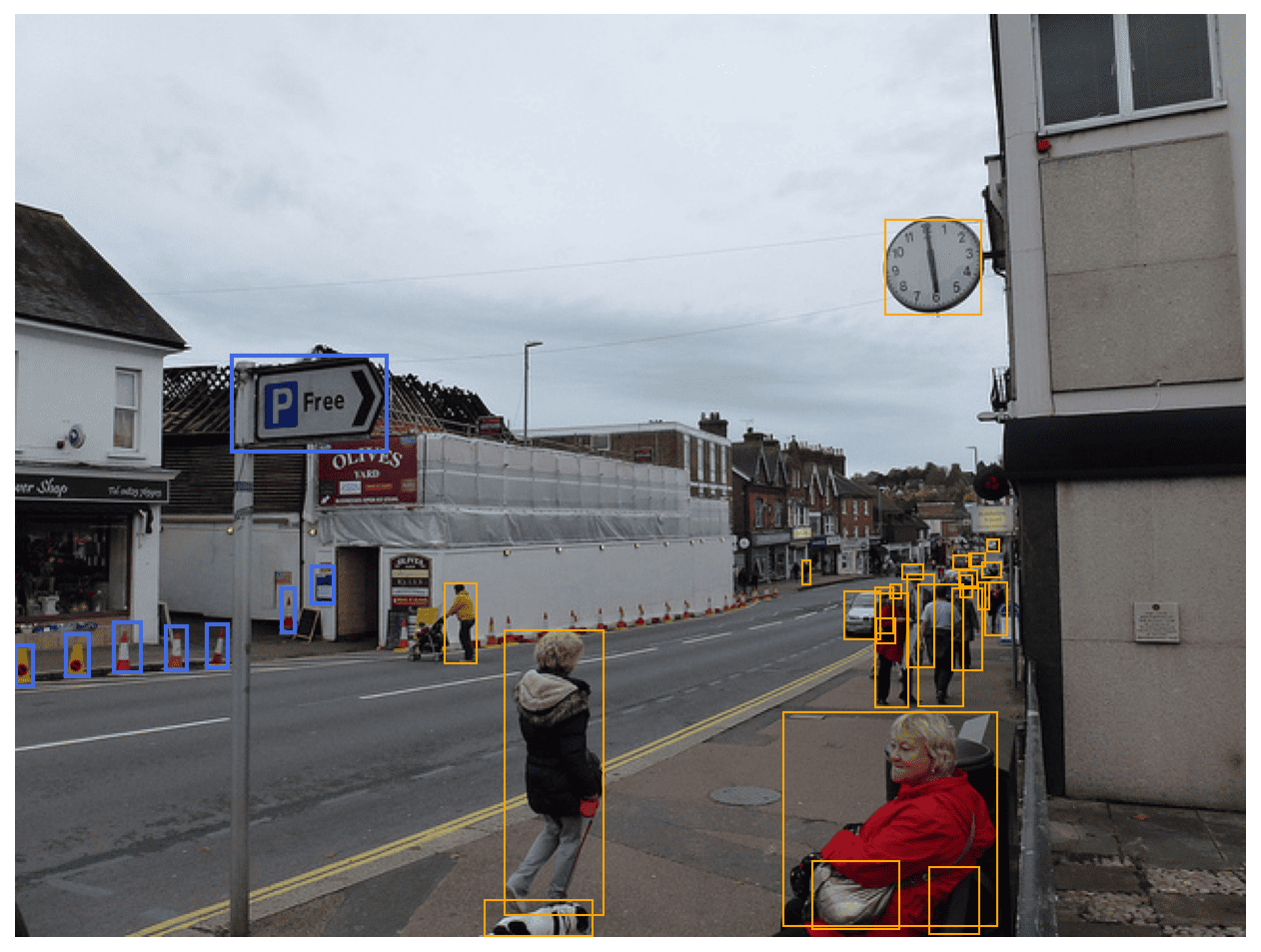} &
    \includegraphics[width=0.48\linewidth]{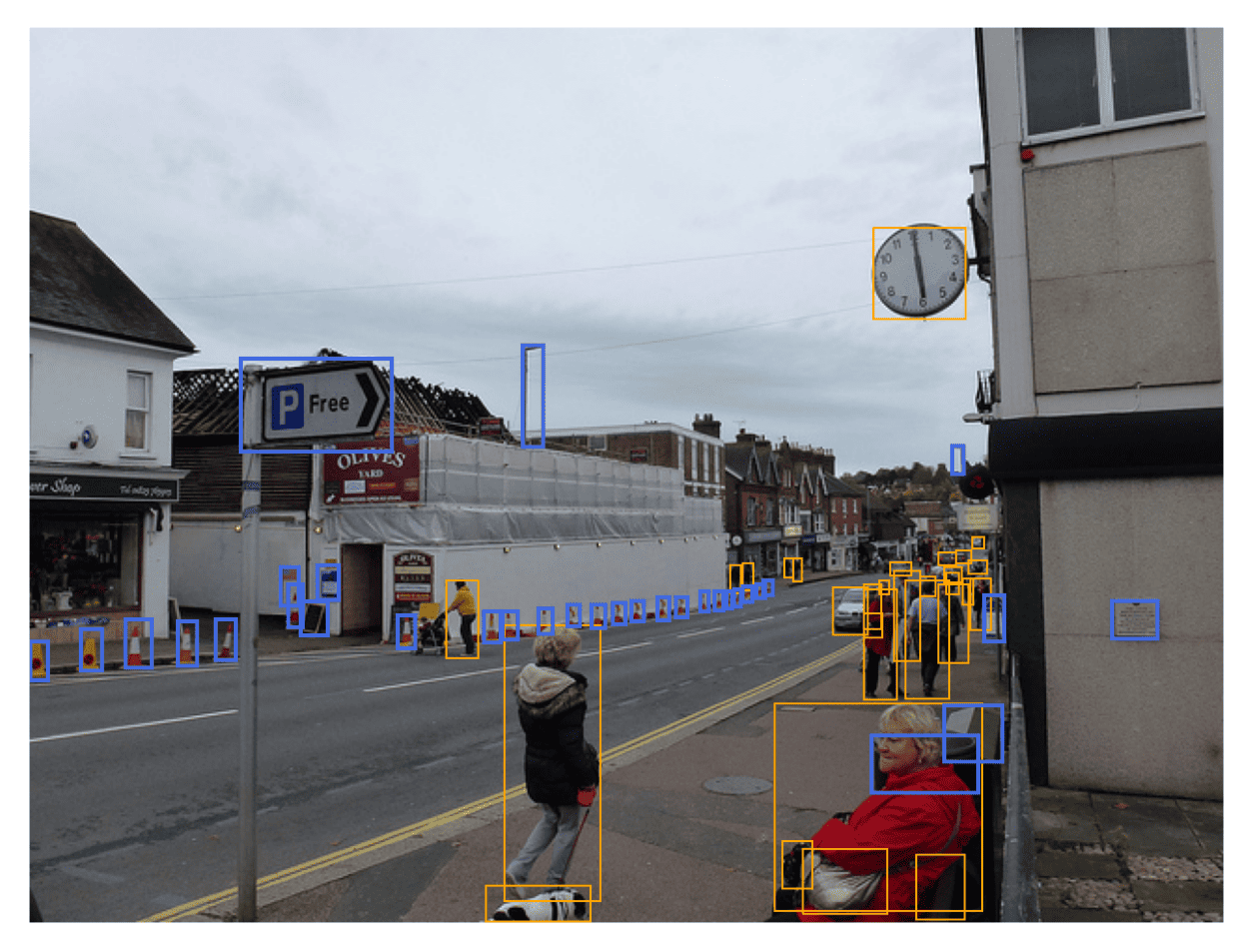} \\
    (c) & (d)
    \end{tabular}%
    }
    \caption{
    \textbf{Examples from COCO-Mixed illustrating distinct limitations (a–c), and an improved annotation from COCO-Open in (d).} 
    \textbf{(a)} Images with composite items (e.g., soup) present ambiguity regarding whether smaller components (e.g., vegetables, meat) should be annotated as distinct objects, making them unsuitable for evaluation. \textbf{(b)} Visual evidence is insufficient to confirm whether the labeled objects are indeed a cat and a bed. For example, the supposed cat may depict another animal or a fur coat. \textbf{(c)} Valid objects such as the traffic cones are missing, which is improved in our proposed COCO-Open dataset as shown in \textbf{(d)}. 
    }
    \label{fig:bad_pictures}
    \end{figure}

    Reliable evaluation of OSOD requires datasets with exhaustive annotations for both known and unknown objects. While COCO-OOD ~\cite{liang2023unknown} and COCO-Mixed ~\cite{liang2023unknown} aim to fulfill these criteria, we observe notable limitations for rigorous OSOD evaluation. Some images in COCO-Mixed introduce ambiguity in defining unknown objects (see Fig.~\ref{fig:bad_pictures}a), and some others are visually inconclusive, making it difficult to assess label correctness (see Fig.~\ref{fig:bad_pictures}b). We therefore believe such images should be removed from OSOD evaluation. Most critically, unknown objects are often not labeled extensively (Fig.~\ref{fig:bad_pictures}c). To address these shortcomings, we introduce a new dataset built upon COCO-Mixed. The 80 original COCO classes \cite{lin2014microsoft} are designated as known, while all other foreground objects are treated as unknown. Annotations were created manually and independently verified by a second annotator to ensure both accuracy and consistency. Notably, our dataset includes over four times more unknown object annotations than original. Additional statistics are provided in Table~\ref{tab:dataset_comparison}.

    \begin{table}[htbp]
    \caption{Comparison of statistics between existing OSOD datasets and our COCO-Open.}
    \begin{center}
    
    \begin{tabular}{|l|c|c|c|}
    \hline
    Test Datasets & \# Images & \# Known & \# Unknown \\
    \hline\hline
    COCO-OOD & 504 & 1222 & 431 \\
    COCO-Mixed & 897 & 4751& 433 \\
    COCO-Open (Ours) & 890  & 4943&1853\\
    \hline
    \end{tabular}
    \end{center}
    
    \label{tab:dataset_comparison}
    \end{table}

	\section{Experiments}
	\label{sec:experiments}
    \subsection{Implementation Details}
    The NAN-SPOT model builds on the pretrained D-DETR checkpoint~\cite{zhu2020deformable}\footnote{Implementation from https://github.com/fundamentalvision/Deformable-DETR.}, which was trained on the full COCO dataset \cite{lin2014microsoft}. On top of this, the objectness module is trained separately using the COCO-OOD dataset 80\% train split~\cite{liang2023unknown}. 
    The random forest uses 100 trees with a maximum depth of 10, requiring at least 10 samples per split and leaf to ensure stable predictions. The MLP is a three-layer network (96–48–24) trained with Adam (learning rate 0.001, batch size 64) using binary cross-entropy loss for object–background classification.
    For evaluation, 3 datasets are considered: COCO-Mixed~\cite{liang2023unknown}, our proposed COCO-Open, and the LVIS dataset~\cite{gupta2019lvis}, which provides additional annotations beyond COCO by introducing many more categories with a long-tailed distribution \cite{gupta2019lvis}. In all cases, the 80 COCO categories are treated as known, while the additional annotations are regarded as unknown objects\footnote{For this reason, we do not adopt benchmarks such as M-OWODB \cite{joseph2021towards} and S-OWODB \cite{gupta2022owdetr}, as they partition COCO categories so that only a subset is known in each task and focus on open-world object detection with incremental learning, which differs from our open-set detection setting.
    }.  %
    At inference, we maintain the original classification scheme for known classes from D-DETR. Namely, predictions with high confidence are treated as known objects, while predictions with low confidence but high objectness are regarded as unknown objects. The confidence threshold is determined using the pretest mode procedure, as detailed in Section~\ref{sec:Determining Inference Threshold in Pretest Mode}. Regions with low objectness are assigned to background. We follow the standard post-process scheme and used top-100 objectness-scoring queries for evaluation. Metrics and baselines evaluated are described in Section~\ref{sec:Metrics and Baselines}.
    
    \subsubsection{Determining Inference Threshold in Pretest Mode}
    \label{sec:Determining Inference Threshold in Pretest Mode}
    Unlike conventional detectors, which rely directly on class-confidence scores derived from classification probabilities, $g(\cdot)$ outputs class-agnostic objectness scores, requiring the calibration of a confidence threshold to separate known detections. Threshold selection is performed in a pretest phase using 20\% COCO-OOD test split, evaluating predictions over the threshold set
    $\boldsymbol{\varepsilon} = [0,\,0.05,\,0.10,\,\dots,\,1]$. For each candidate $\varepsilon_i$ we compute a combined metric $S(\varepsilon_i)$:
    \begin{equation}
    S(\varepsilon_i) \;=\; 
    \frac{\text{mAP}_{\text{known}}(\varepsilon_i)}{\max_j \text{mAP}_{\text{known}}(\varepsilon_j)}
    \;+\;
    \frac{R_{u}(\varepsilon_i)}{\max_j R_{u}(\varepsilon_j)},
    \end{equation}
    and select $\varepsilon^{*} = \arg\max_i S(\varepsilon_i)$.

    \subsubsection{Metrics and Baselines}
    \label{sec:Metrics and Baselines}
    To ensure comparability, we evaluate NAN-SPOT against four D-DETR-based OWOD baselines: OW-DETR~\cite{gupta2022owdetr}, CAT~\cite{ma2023cat}, PROB~\cite{zohar2023prob} and HYP-OW~\cite{doan2024hyp}. %
    Following evaluation metrics are employed:  
    (1) mAP, the mean average precision for known classes, averaged over Intersection over Union (IoU) thresholds in $[0.5 : 0.05 : 0.95]$;  
    (2) R\textsubscript{u} and AP\textsubscript{u}, recall and average precision of unknown objects over the same IoU range ;  
    (3) Wilderness Impact (WI)~\cite{dhamija2020overlooked} measures the impact of unknown objects on the detector’s performance at the recall level of 0.8~\cite{han2022expandinglowdensity,joseph2021towards}. It is computed as: $WI = \tfrac{FP_u}{TP_k + FP_k}$, where $TP_k$ and $FP_k$ denote true and false positives for known classes, and $FP_u$ represents false positives for unknowns;
    (4) AOSE (Absolute Open-Set Error)~\cite{miller2018dropout}, which counts the total number of unknown objects misclassified as any known class.
    
    \subsection{Main Results}
    
    \subsubsection{Quantitative Analysis} 
    Table~\ref{tab:main_results} compares NAN-SPOT with OW-DETR~\cite{gupta2022owdetr}, CAT~\cite{ma2023cat}, PROB~\cite{zohar2023prob} and HYP-OW~\cite{doan2024hyp} on the COCO-Mixed \cite{liang2023unknown}, COCO-Open and LVIS dataset \cite{gupta2019lvis}. 
    \begin{table*}[htbp!]
    \caption{
    \textbf{Comparison with baseline methods on COCO-Mixed, COCO-Open and LVIS dataset.}
    \normalfont NAN-SPOT achieves the best balance on unknown recall and average precision, with competitive WI and AOSE, demonstrating strong unknown object detection performance with minimal impact on known-class detection. The consistency of performance trends across COCO-Mixed, COCO-Open and LVIS indicates that the improvements of our model are robust. Best results are in \textbf{bold}, second best are \underline{underlined}. 
    }
    \begin{center}
    \centering
    \begin{threeparttable}
    \begin{minipage}{0.75\textwidth} 
    \begin{tabular}{|l|l|c|c|c|c|c|}
    \hline
    \textbf{Dataset} & Model & mAP ($\uparrow$) & R$_{u}$ ($\uparrow$) & WI ($\downarrow$) & AOSE ($\downarrow$) & \makecell{Training \\ Time (h) \tnote{a}} \\
    \hline \hline
    \multirow{6}{*}{COCO-Mixed~\cite{liang2023unknown}} 
    & OW-DETR \cite{gupta2022owdetr} & 0.468 & 0.178  & 2.33 & 1296 & 323 \\
    & CAT \cite{ma2023cat} & \textbf{0.505} & 0.159 & 2.29 & 1343& 476 \\
    & PROB \cite{zohar2023prob} & 0.495 & 0.469 & 2.15 & 770 & 140 \\
    & HYP-OW \cite{doan2024hyp}  & 0.501 & \textbf{0.487} & 2.18 & 326 & 142 \\
    & Ours: NAN-SPOT-mlp & \textbf{0.505} & 0.470& \textbf{1.65} & \textbf{104} & \underline{0.03} \\
    & Ours: NAN-SPOT-rf & \textbf{0.505} & \underline{0.481}& \textbf{1.65} & \textbf{104} & \textbf{0.02} \\
    \hline
    \multirow{6}{*}{COCO-Open} 
    & OW-DETR \cite{gupta2022owdetr} & 0.457 & 0.135 & 7.42 & 3755 & 323\\
    & CAT \cite{ma2023cat} & \textbf{0.499} & 0.096 & 7.32 & 4131 &  476 \\
    & PROB \cite{zohar2023prob} & 0.484 & 0.376 & 7.03 & 1855 & 140 \\
    & HYP-OW \cite{doan2024hyp} & 0.486 & 0.375 & 5.96 & 881 & 142 \\
    & Ours: NAN-SPOT-mlp & \underline{0.494} & \underline{0.388} & \textbf{4.88}& \textbf{308} & \underline{0.03}\\
    & Ours: NAN-SPOT-rf & 0.493 & \textbf{0.392} & \textbf{4.88}& \textbf{308} & \textbf{0.02}\\
    \hline
    \multirow{6}{*}{LVIS~\cite{gupta2019lvis}} 
    & OW-DETR \cite{gupta2022owdetr} & 0.241 & 0.047 & \underline{8.44} & 25516 & 323\\
    & CAT \cite{ma2023cat} & 0.267 & 0.035 & \textbf{8.05} & 25560 & 476 \\
    & PROB \cite{zohar2023prob} & 0.251 & \underline{0.177} & 8.87 & 11848 & 140 \\
    & HYP-OW \cite{doan2024hyp} & 0.254 & 0.174 & 9.73 & 6348 & 142 \\
    & Ours: NAN-SPOT-mlp & \textbf{0.272} & 0.172 & 9.31& \underline{4749} & \underline{0.03}\\
    & Ours: NAN-SPOT-rf & \underline{0.271} & \textbf{0.181} & 9.36 & \textbf{4640} & \textbf{0.02}\\
    \hline
    \end{tabular}%
    \begin{tablenotes}
    \footnotesize
    \item[a] OW-DETR, CAT, PROB and HYP-OW were trained on the COCO dataset \cite{lin2014microsoft} using NVIDIA H100 GPUs (94 GB), with reported training times in GPU hours. NAN-SPOT, in contrast, was trained on NVIDIA GeForce RTX 4090 GPU (24 GB).
    \end{tablenotes}
    \end{minipage}
    \end{threeparttable}
    \end{center}
    \label{tab:main_results}
    \end{table*}%
    Among these benchmarks, COCO-Open poses the most demanding evaluation, as it provides exhaustive annotations for both known and unknown objects. This results in lower absolute performance across models compared to COCO-Mixed, where annotations are less complete but still of reasonable quality for known categories. In contrast, although LVIS provides a large number of annotations, they are incomplete at the per-category level, with many instances of defined classes unlabeled. Furthermore, object definitions are sometimes inconsistent across categories. As a result, correct detections on unlabeled instances are counted as false positives, artificially depressing recall and inflating error metrics. Thus, the much lower performance observed on LVIS reflects annotation limitations rather than inherently greater task difficulty.

    Across all datasets, consistent trends emerge among existing baselines. CAT achieves relatively high mAP, reflecting strong closed-set detection, but performs poorly on open-set metrics. Its low unknown recall and high WI and AOSE indicate frequent misclassifications of unknowns as known. PROB takes the opposite trade-off, achieving higher unknown recall and substantially reducing WI and AOSE relative to CAT and OW-DETR, though at the cost of slightly reduced mAP. HYP-OW shows competitive unknown recall on COCO-Mixed but lacks robustness on more challenging benchmarks. In contrast, NAN-SPOT consistently achieves the most favorable balance between known-class accuracy and open-set robustness. On COCO-Mixed, it matches the best mAP while dramatically lowering WI and AOSE compared to all baselines. On COCO-Open, it attains the highest unknown recall (39.2\%) together with the lowest WI (4.88) and AOSE (308), demonstrating its ability to detect unknown objects without introducing confusion to known-unknown separation. On LVIS, NAN-SPOT again attains the highest unknown recall (18.1\%), while remaining competitive in mAP.
    
    Overall, NAN-SPOT demonstrates robust performance across all datasets, consistently delivering competitive mAP together with leading open-set metrics. Crucially, these gains are accomplished with minimal computational cost: whereas OW-DETR, CAT, PROB and HYP-OW require hundreds of GPU-hours, NAN-SPOT can be trained within minutes from a D-DETR checkpoint. This combination of accuracy, robustness, and efficiency establishes NAN-SPOT as the most practical solution for open-set detection across diverse benchmarks.

    \subsubsection{Qualitative Analysis}

    \begin{figure*}[h!]
    \begin{center}
    \includegraphics[width = \linewidth]{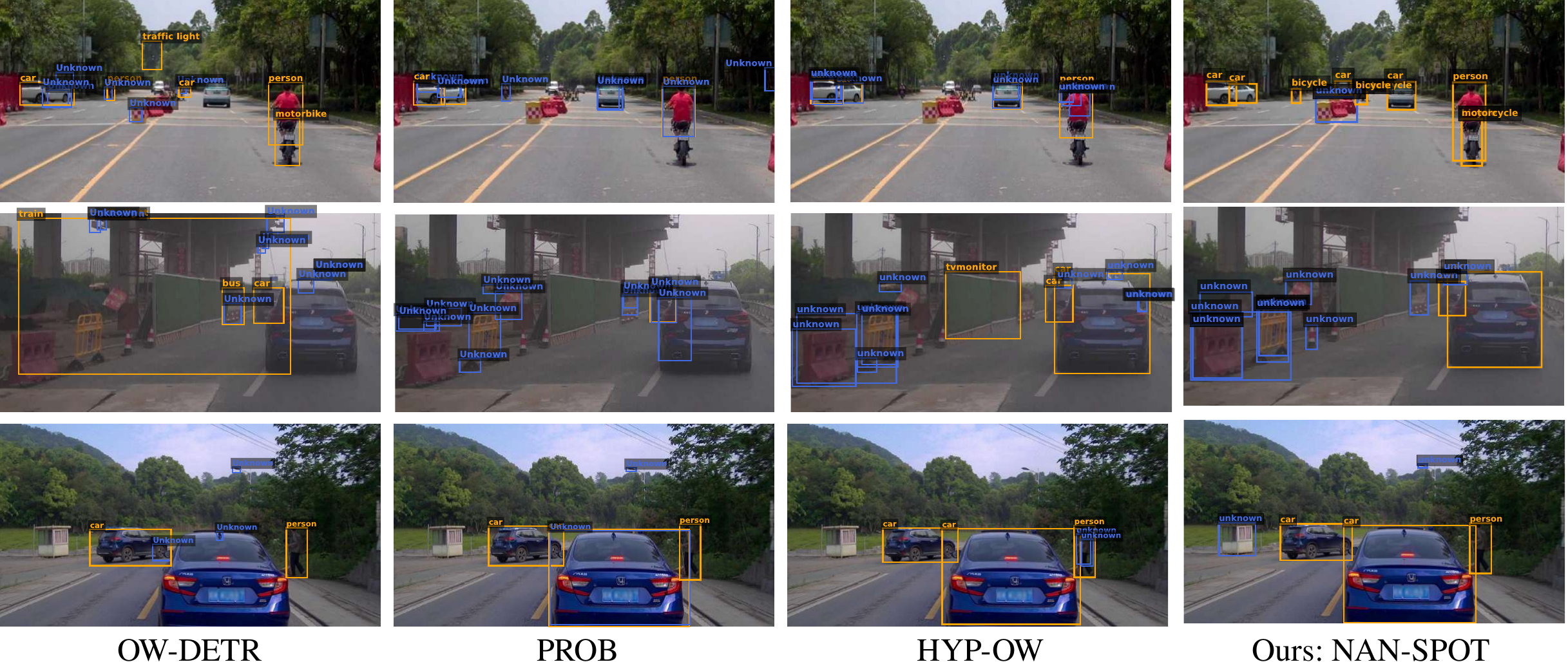}
    \end{center}
       \vspace{-10pt}
       \caption{
       \textbf{Qualitative results on CODA.}
    Comparison of OW-DETR~\cite{gupta2022owdetr}, PROB~\cite{zohar2023prob}, HYP-OW~\cite{doan2024hyp} and NAN-SPOT on detecting \textcolor{known}{known} and \textcolor{unknown}{unknown} objects. Same number of top-$k$ predictions are shown per image for fair comparison.}
    \label{fig:result}
    \end{figure*}%

    Fig.~\ref{fig:result} presents qualitative comparisons across OW-DETR~\cite{gupta2022owdetr}, PROB~\cite{zohar2023prob}, HYP-OW~\cite{doan2024hyp}, and our NAN-SPOT using the same top-$k$ predictions on the CODA dataset~\cite{li2022coda}. The examples highlight distinct failure modes and strengths across different methods.
    OW-DETR, as one of the earliest OWOD approaches, exhibits limited robustness in handling unknown objects. It frequently fails to detect unknown instances, such as the traffic barriers in the second example, and tends to hallucinate about known categories, producing false positives like the traffic light in the first image and the train in the second image.
   PROB demonstrates the opposite tendency, often suppressing known objects. Its scoring mechanism strongly favors unknown predictions, resulting in very few known objects appearing in the top-$k$ visualizations. As a consequence, clearly visible cars in both the first and second images are missed.
    HYP-OW shows inconsistent localization quality and frequently generates overlapping bounding boxes, as seen for the middle green car in the first image and the person in the third image.
    In contrast, NAN-SPOT provides a more balanced detection of both known and unknown objects. It consistently detects key known objects such as cars and persons, while simultaneously highlighting plausible unknown objects without excessive false positives. Notably, NAN-SPOT identifies road barriers in the first and second images and the small roadside structure in the third image as unknown, demonstrating strong generalization to diverse unseen object types. Overall, its predictions are clean, well-localized, and semantically coherent, supporting its strong quantitative results.

    \subsubsection{Ablation Studies} 
    Table~\ref{tab:ablation} presents the impact of removing individual components of NAN-SPOT on COCO-Open. The confidence has the strongest influence on both mAP and unknown recall, which is expected since it is the only directly trained signal; removing it leads to a marked drop in performance across both metrics. The box feature contributes primarily to stabilizing objectness estimates, as its removal has a smaller but still noticeable effect on recall. Finally, the addition of NAN yields a clear performance gain, significantly increasing unknown recall (from 0.374 to 0.392). This demonstrates that NAN complements the existing features by providing a robust signal for distinguishing unknown objects, leading to more effective open-set detection. 
    
    \begin{table}[htbp]
    \caption{
    \textbf{Ablation study of NAN-SPOT-rf on the COCO-Open dataset.}
    \normalfont Removing the score feature leads to the largest drop in performance, confirming its vital role as the trained signal. Box features have smaller but consistent contributions. Excluding NAN leaves mAP unchanged but lowers unknown recall, demonstrating its importance for distinguishing unknown objects.
    }
    \begin{center}
    \begin{threeparttable}
    \begin{tabular}{|l|c|c|}
    \hline
     \textbf{Model} & mAP ($\uparrow$) & R$_{u}$ ($\uparrow$)     \\
    \hline \hline
    w/o NAN $f_{\text{NAN}}$   & 0.493 & 0.374    \\
    w/o box center to image edge $d_{\text{center}}$   & \textbf{0.494} & 0.385    \\
    w/o box edge to image edge $d_{\text{edge}}$  & 0.493 & 0.389    \\
    w/o box area $s_{\text{box}}$  & 0.493 & 0.383    \\
    w/o confidence score $p_{\text{conf}}$ & 0.470 & 0.347     \\
    NAN-SPOT (ours)  & 0.493 & \textbf{0.392}    \\
    \hline
    \end{tabular}%
    \end{threeparttable}
    \end{center}
    \label{tab:ablation}
    \end{table}
    
    \subsubsection{Effectiveness of Pretest Threshold Determination}
    
\begin{figure}[t]
    \begin{center}
    \includegraphics[width=0.7\linewidth]{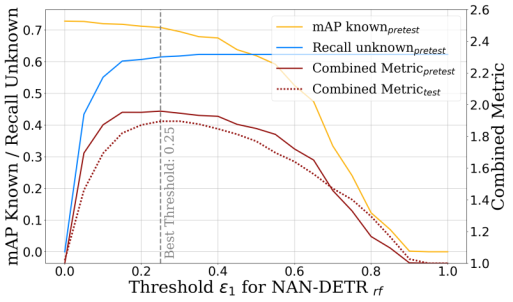} 
    \end{center}
       \caption{\textbf{Threshold calibration via pretest data.} Performance of NAN-SPOT across thresholds $[0:0.05:1]$ for \textcolor{known}{known mAP}, \textcolor{unknown}{unknown recall}, and the \textcolor{darkred}{combined metric}. 
    The pretest procedure on COCO-OOD pretest data identifies the optimal threshold at ${\varepsilon_{1}}^*=0.25$, which also generalizes well to test data (COCO-Open). }
    \label{fig:pretest_threshold}
    \end{figure}

    The pretest procedure on the COCO-OOD test-split identifies an optimal threshold of ${\varepsilon_{1} }^*= 0.25$. As shown in Fig.~\ref{fig:pretest_threshold}, known mAP decreases monotonically with increasing threshold values, while unknown recall monotonically increases. The combined metric, designed to balance known-class accuracy and unknown recall, peaks at ${\varepsilon_{1}}^*$. When the same analysis is repeated on the COCO-Open test dataset, a nearly identical trend is observed, with the combined metric again reaching its maximum near the same threshold. These consistent results confirm that the pretest mode provides a reliable means of threshold calibration for open-set object detection, while relying only on  pre-test data and avoiding any risk of test-data leakage.

	\section{Summary and Conclusions}
	\label{sec:conclusion}
	In this work, we introduce NAN-SPOT, a training-light framework that leverages the NAN metric to estimate objectness, enabling effective separation of both known and unknown objects from background without retraining the base detector. To facilitate more rigorous evaluation, we also propose COCO-Open, a densely annotated dataset with significantly expanded unknown object labels. Our experiments show that NAN-SPOT offers enhanced performance on unknown object detection compared to other methods, while maintaining competitive performance on known objects. We find the straightforward yet effective design of NAN-SPOT, grounded in the NAN metric from the hidden layer, to be particularly compelling. The findings underscore the promising role of hidden representations in modern large models, presenting opportunities for future research to extract and apply this information in advancing open-world perception and beyond.
	
	\section*{ACKNOWLEDGMENTS}
    This work was funded by the Förderverein Mobile Arbeitsmaschinen e.V. (MOBIMA) within the research project “Sichere Objektklassifizierung” (MOBIMA-Nr. 107 I).
    
	The authors also gratefully acknowledge the scientific support and resources of the AI service infrastructure LRZ AI Systems provided by the Leibniz Supercomputing Centre (LRZ) of the Bavarian Academy of Sciences and Humanities (BAdW), funded by Bayerisches Staatsministerium für Wissenschaft und Kunst (StMWK).
	
	\bibliographystyle{IEEEtran}
	\bibliography{root} 
	
\end{document}